\documentclass{article}

\usepackage{microtype}
\usepackage{graphicx}
\usepackage{booktabs} 

\usepackage{xspace}
\usepackage{glossaries}
\usepackage[inline]{enumitem}
\usepackage{subfig}
\usepackage[mode=buildnew]{standalone}
\usepackage{pgf,tikz}
\usepackage{endnotes}

\definecolor{bluekeywords}{rgb}{0.13, 0.13, 1}
\definecolor{greencomments}{rgb}{0, 0.5, 0}
\definecolor{redstrings}{rgb}{0.9, 0, 0}
\definecolor{graynumbers}{rgb}{0.5, 0.5, 0.5}

\usepackage{listings}
\usepackage{hyperref}

\usepackage[accepted]{sysml2019}

\sysmltitlerunning{Addressing Algorithmic Bottlenecks in Elastic Machine Learning with Chicle}

\definecolor{green}  {rgb} {0.31,0.60,0.02} 
\definecolor{blue}   {rgb} {0.13,0.29,0.53} 
\definecolor{red}    {rgb} {0.64,0.00,0.00} 

\newcommand{\sys}{Chicle\xspace}
\newcommand{\Sys}{Chicle\xspace}
\newcommand{\concept}{uni-tasks\xspace}
\newcommand{\Concept}{Uni-tasks\xspace}
\newcommand{\Sec}[1]{\S{\ref{sec:#1}}}
\newcommand{\Fig}[1]{Figure~{\ref{fig:#1}}}
\newcommand{\Figs}[2]{Figures~{\ref{fig:#1} and \ref{fig:#2}}}

\newcommand{\Tab}[1]{Table~{\ref{tab:#1}}}
\newcommand{\Eq}[1]{Equation~{\ref{eq:#1}}}
\newcommand{\Lst}[1]{Listing~{\ref{lst:#1}}}

\newcommand{\note}[1]{} 

\newcommand{\todo}[1]{\note{TODO: #1}}
\newcommand{\KOU}[1]{\note{KOU: #1}}

\newcommand{\ok}[1]{#1}

\begin{document}
\newacronym{cocoa}{CoCoA}{Communication-efficient distributed dual Coordinate Ascent}
\newacronym{sgd}{SGD}{stochastic gradient descent}
\newacronym{sdca}{SDCA}{stochastic dual coordinate ascent}
\newacronym{scd}{SCD}{stochastic coordinate descent}
\newacronym{svm}{SVM}{support vector machine}
\newacronym{dg}{DG}{duality-gap}
\newacronym{rdma}{RDMA}{remote direct memory access}
\newacronym{rpc}{RPC}{remote procedure call}
\newacronym{glm}{GLM}{generalized linear model}
\newacronym{nn}{NN}{neural network}
\newacronym{dnn}{DNN}{deep neural network}
\newacronym{rnn}{RNN}{recurrent neural network}
\newacronym{cnn}{CNN}{convolutional neural network}
\newacronym{ml}{ML}{machine learning}
\newacronym{bsp}{BSP}{bulk synchronous parallel}
\newacronym{ssp}{SSP}{stale synchronous parallel}
\newacronym{scm}{SCM}{storage class memory}
\newacronym{sgd}{SGD}{Stochastic Gradient Descent}
\newacronym{msgd}{mSGD}{Mini-batch SGD}
\newacronym{lsgd}{lSGD}{Local SGD}
\newacronym{ps}{PS}{Parameter Server}

\twocolumn[
\sysmltitle{Addressing Algorithmic Bottlenecks in Elastic Machine Learning with Chicle}



\sysmlsetsymbol{equal}{*}

\begin{sysmlauthorlist}
\sysmlauthor{Michael Kaufmann}{ibm,kit}
\sysmlauthor{Kornilios Kourtis}{ibm}
\sysmlauthor{Celestine Mendler-D\"unner}{ucb}
\sysmlauthor{Adrian Sch\"upbach}{ibm}
\sysmlauthor{Thomas Parnell}{ibm}
\end{sysmlauthorlist}

\sysmlaffiliation{ibm}{IBM Research, Zurich, Switzerland}
\sysmlaffiliation{kit}{Karlsruhe Institute of Technology, Karlsruhe, Germany}
\sysmlaffiliation{ucb}{UC Berkeley, work conducted while at IBM Research}

\sysmlcorrespondingauthor{Michael Kaufmann}{kau@zurich.ibm.com}

\sysmlkeywords{Machine Learning, SysML}

\vskip 0.3in

\begin{abstract}

Distributed machine learning training is one of the most common and important
workloads running on data centers today, but it is rarely executed alone.
Instead, to reduce costs, computing resources are consolidated and shared by
different applications. In this scenario, elasticity and proper load balancing
are vital to maximize efficiency, fairness, and utilization. Currently,
most distributed training frameworks do not support the aforementioned
properties. A few exceptions that do support elasticity, imitate
generic distributed frameworks and use micro-tasks.

In this paper we illustrate that micro-tasks are problematic for machine
learning applications, because they require a high degree of parallelism which
hinders the convergence of distributed training at a pure algorithmic level
(i.e., ignoring overheads and scalability limitations).
To address this, we propose \Sys, a new elastic distributed training framework
which exploits the nature of machine learning algorithms to implement elasticity
and load balancing without micro-tasks.  We use \Sys to train deep neural
network as well as generalized linear models, and show that \Sys achieves
performance competitive with state of the art rigid frameworks, while
efficiently enabling elastic execution and dynamic load balancing.

\end{abstract}
]



\printAffiliationsAndNotice{}  

\section{Introduction}
\label{sec:introduction}
\glsresetall

The ever-growing amounts of data are fueling impressive advances in \gls{ml}, but
depend on substantial computational power to train the corresponding models. As
a result, many research works focus on addressing \emph{scalability} of
distributed training across multiple machines. State of the art algorithms include
\gls{msgd}~\cite{robbins1951, kiefer1952, rumelhart1988} and
\gls{lsgd}~\cite{lin2018} for \glspl{dnn} as well as
\gls{cocoa}~\cite{jaggi2014, smith2018} for \glspl{glm}.

Less work, however, has focused on \emph{efficiency}, which is equally (if not
more) important because it effectively provides more computational power at the
same cost. Indeed, most works on distributed \gls{ml} assume that they can
operate on dedicated clusters, which is rarely the case in practice where
\gls{ml} applications co-inhabit common
infrastructure with other applications.
In these shared environments, efficiency depends on two properties:
\emph{elastic execution}: dynamically adjusting resource (e.g., CPUs, GPUs,
nodes) usage as their availability changes, and \emph{load balancing}:
distributing workload across heterogeneous resources~\cite{ou2012,
  delimitrou2014} such that faster resources do not have to wait for slower
ones.
Elastic execution, specifically, enables optimization opportunities for \gls{ml}
applications where scaling-in or -out as training progresses can increase
accuracy and reduce training time \cite{kaufmann2018}.

As of today, most \gls{ml} distributed frameworks (e.g., \citet{abadi2016,
  paszke2017}) do not support elastic execution nor load balancing, which makes
them inherently inefficient in shared environments and on heterogeneous
clusters.  Recently, recognizing the importance of elasticity, a number of
systems attempt to address elasticity~\cite{zhang2017, harlap2017, qiao2018} for
\gls{ml} applications using micro-tasks or similar mechanisms. Micro-tasks,
where work is split up into a large number of short tasks executed as resources
become available, have been extensively used in generic distributed application
frameworks to address elasticity and load balancing~\cite{zaharia2010,
  ousterhout2013}, so they seem a natural fit for this problem.

In this paper we argue that micro-tasks are ill-suited for ML training because
they require a large number of short independent tasks for efficient scheduling.
In order to support full system utilization, the number of tasks has to be
chosen based on the largest possible degree of parallelism an elastic system
could potentially experience.  The number of tasks, in turn, constitutes a lower
bound on the data parallelism of each update which means that you need to pick
the mini-batch size in \gls{msgd}\footnote{The mini-batch size needs to be
  chosen as a multiple of the number of tasks in order to keep relative job
  overheads low.} or the number of partitions in CoCoA accordingly.  This
however is not a desirable thing to do from an algorithmic point of view, since
it is widely acknowledged that data parallelism comes at the cost of convergence
in distributed \gls{ml} applications.  Note that when talking about convergence
we refer to epochs to converge, where an epoch refers to one pass through the
entire dataset.

\begin{figure}[t]
  \centering
  \subfloat[CIFAR-10 (\glstext{msgd}/\glstext{cnn})] {
    \centering \includegraphics[page=1,scale=0.55]{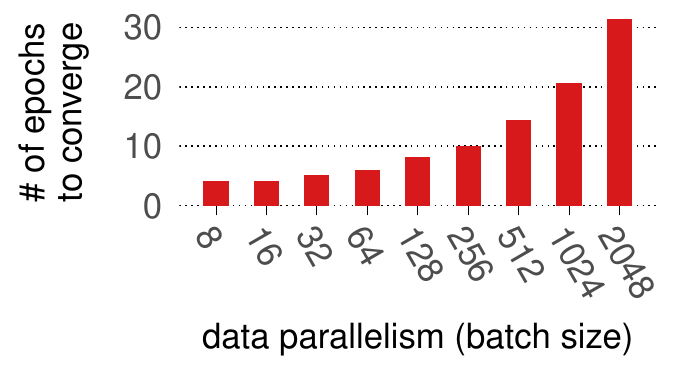}
    \label{fig:introduction:correlation-bs-epochs:msgd}
  }
  ~~
  \subfloat[Criteo (\glstext{cocoa}/\glstext{svm})] {
    \centering \includegraphics[page=2,scale=0.55]{figures/batch_size_vs_num_epochs.pdf}
    \label{fig:introduction:correlation-bs-epochs:cocoa}
  }
  \caption{Example of the correlation between data parallelism and the number of
    epochs needed to achieve a certain training goal.}
  \label{fig:introduction:correlation-bs-epochs}
\end{figure}

Extensive studies of this impact for \gls{msgd} have, among others, been
conducted by \citet{shallue2019}, \citet{keskar2016} and \citet{goyal2017}.
\Figs{introduction:correlation-bs-epochs:msgd}
{introduction:correlation-bs-epochs:cocoa} also exemplify this.
The training of a simple \gls{cnn} on the CIFAR-10 dataset using \gls{msgd}
requires 44\% more epochs to converge when increasing the batch size from 256 to
512.  Similarly, doubling the number of partitions from 16 to 32 for the
training of a \note{\gls{svm}} on the Criteo dataset using
\gls{cocoa}~\cite{jaggi2014, smith2018} increases the number of epochs to
converge by 65\%.
While mitigation strategies, such as warm-up~\cite{goyal2017} and layer-wise
adaptive rate scaling~\cite{you2017large} exist, the fundamental problem
remains.  \KOU{After some, hopefully convincing details, we 've reached our
  conclusion:} Overall, micro-tasks lead to an inherent conflict between the
number of tasks to use for scheduling efficiency, where higher is better, and
algorithmic \gls{ml} training efficiency, where lower is better.

Fortunately, as we show in this paper, the iterative nature of \gls{ml}
applications allows implementing load balancing and elasticity without
micro-tasks, thus eliminating the above inherent conflict.
We realize our ideas
in \Sys\footnote{\Sys is the Mexican-Spanish word for latex from the sapodilla
tree that is used as basis for chewing gum and a reference to \Sys's
elasticity.}, an elastic, load balancing distributed framework for
iterative-convergent \gls{ml} training applications.  \Sys combines scheduling
flexibility with the efficiency of special-purpose \emph{rigid} \gls{ml}
training frameworks. \Sys uses \concept and schedules (stateful) data chunks
instead of tasks.
Each node executes only a single (multi-threaded) task that
processes training samples from multiple data chunks within a single execution
context.  Data chunks can be moved efficiently between tasks to balance load and
to scale in and out.
This allows \Sys to use the optimal level of data parallelism for the currently
used number of resources and combines scheduling with algorithmic efficiency.
Conversely, \Sys is able to efficiently adjust the resource allocation based on
feedback from the training algorithm and resource availability.
%
%
%
The main contributions of our work are:
\begin{enumerate}[label={\bfseries\arabic*)}]
%
\item We propose \emph{\concept}, a new task model that removes the conflict
  between scheduling and algorithmic efficiency.  We implement a prototype
  thereof in \Sys, a distributed \gls{ml} framework that enables elastic
  training and dynamic load balancing in heterogeneous clusters.

\item Our evaluation illustrates that \concept require significantly fewer
  epochs, and subsequently less time, to converge in elastic and load-balancing
  scenarios compared to micro-tasks.

\end{enumerate}

Our paper is structured as follows: First, we provide necessary background
information on the relationship between data parallelism and convergence for
\gls{ml} training algorithms as well as requirements for elastic execution in
\Sec{background} followed by a discussion of the main ideas behind \concept
(\Sec{concept}).  We continue with a detailed description of \Sys's design and
implementation (\Sec{system}) and present results of our experimental evaluation
(\Sec{evaluation}) and conclude (\Sec{conclusion}).

\section{Background \& motivation}
\label{sec:background}



Increasing parallelism for distributed execution of \gls{ml} training workloads has
well-understood tradeoffs. On one hand, ample parallelism results in less work per each
independent execution unit (\emph{task}) which leads to increased
overheads~\cite{totoni2017}.  On the other hand, ample parallelism allows utilizing many
nodes and enables efficient scheduling~\cite{ousterhout2013}, dealing with load
imbalances, and supporting elasticity.
Elasticity specifically is increasingly important, since to maximize the efficiency,
distributed applications are expected to scale-in and -out based on workload demands of
themselves and their cohabitants.
Indeed, exposing ample parallelism by dividing the problem into many micro-tasks is the
standard way to implement elasticity despite the resulting execution overheads.
Litz~\cite{qiao2018}, for example, a recent \gls{ml} elastic framework uses micro-tasks
and reports up to 23\% of execution overhead.

While these overheads are important, our work is motivated by another tradeoff that is
specific to \gls{ml} applications but not well recognized in the \gls{ml} systems
community: increased data parallelism hinders the convergence of \gls{ml} training.  In
contrast to overheads, this problem exists purely at the algorithmic level. Generally,
distributed training algorithms require more steps to converge in the face of high
parallelism~\cite{shallue2019}. The implication for building elastic \gls{ml} frameworks
is that using micro-tasks, i.e., ample parallelism to gain scheduling flexibility, leads
to an inherent trade-off in terms of the number of the examples that need to be processed
to converge to a solution.

In this section, we motivate our design by illustrating this issue in two different ML
algorithms: Mini-batch \gls{sgd}, extensively used to train neural networks, and
\gls{cocoa}, a state-of-the art framework for distributed training of \glspl{glm}. Prior
to that, we provide some necessary background on elastic scheduling and \gls{ml} training.

\subsection{Elasticity and load balancing}
\label{sec:background:elasticity_load_balancing}

Both \emph{load balancing} and elasticity are necessary to efficiently utilize
shared infrastructure.
Both are typically implemented using micro-tasks in generic analytics
frameworks, such as Spark~\cite{zaharia2010} and \gls{ml}
frameworks~\cite{qiao2018, zhang2017}, where work is divided into a large number
of tasks that are distributed among nodes.
Tasks, i.e., self-contained, atomic entities of a function and input data, are a
common abstraction of work, and represent the scheduling unit.

Under a task scheduling system, a large number of tasks are required to achieve
efficiency.  To allow elastic scale-out during training, the number of tasks
needs to be at least as large as the maximum number of nodes that will be
available at any point during training. Furthermore, common practice
over-provisions nodes with many tasks per node to allow for efficient load
balancing.  The Spark tuning guidelines~\cite{spark_tuning_guide}, for instance,
recommend to use of up to 2--3 tasks per available CPU, while other works
propose using millions of tasks~\cite{ousterhout2013}.

%
%

\todo{figure that exemplifies why we need many tasks for load balancing by
  showing that if we only have N tasks per node, we can only really balance node
  performances that differ by a factor of 1/N.??}





\subsection{Distributed training algorithms}
\label{sec:background:mlprimer}

Next, we discuss training in general and introduce two training algorithms that
we use in this paper.
Most distributed training algorithms iteratively refine a model $\mathbf m$ on a
training dataset $D$ such that $ \mathbf m$ converges towards a state that
minimizes or maximizes an objective function.  During each iteration $i$, an
updated model $\mathbf m^{(i)}$ is computed on a randomly chosen subset
$\widehat{D} \subseteq D$:
\begin{equation} \label{eq:background:modelupdate}
  \mathbf m^{(i)} =\mathbf m^{(i-1)} + f_{\Delta}(\mathbf m^{(i-1)}, \widehat{D})
\end{equation}

The update function $f_{\Delta}$ is computed in a data parallel manner across
$K$ nodes by splitting up $\widehat{D}$ into $K$ disjoint partitions $D_{k}
\subseteq \widehat{D}$. 

\begin{equation} \label{eq:background:updatecomputation}
  f_{\Delta}(\mathbf m^{(i-1)},\widehat{D}) = {{1}\over{K}} \sum_{k=1}^{K}
  f_{\Delta,k}(\mathbf m^{(i-1)},D_{k})
\end{equation}

\begin{figure}[h!]
  \centering
  \includegraphics[width=0.9\linewidth]{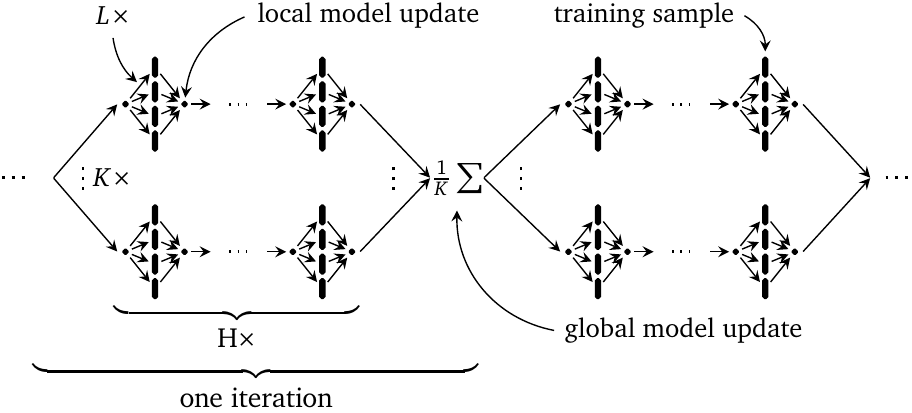}
  \caption{General structure of distributed \gls{ml} algorithms we consider in
    this paper.}
  \label{fig:background:algostructure}
\end{figure}

The computation of $f_{\Delta}$ is self-correcting to a certain degree, i.e.,
bounded errors are averaged out in subsequent iterations, and can therefore be
tolerated.  This property is often exploited for \gls{ml}-specific
optimizations, e.g., to mitigate stragglers~\cite{cipar2013, cui2014, dutta2018,
  ho2013}.
The general structure of the algorithms we are considering is depicted in
\Fig{background:algostructure}: $K$ workers independently work on separate
subproblems $f_{\Delta,k}$, each defined on a different partition $D_k$ of the
data and then combine their results to update a global model $\mathbf{m}$, which
forms the basis of the next iteration.  During each iteration, a worker
processes $H \times L$ samples, of which $H$ different sets of $L$ independent
samples are processed sequentially.  After each set of $L$ samples, a local
model update is performed, such that learning on subsequent samples within an
iteration can exploit knowledge gained so far.  

While our approach is applicable to a wide set of distributed \gls{ml} training
algorithms, in this paper, we focus on the following two algorithms.

\paragraph*{Local \gls{sgd}~\cite{lin2018}.}
\label{sec:background:mlprimrt:msgd}
A state-of-the-art algorithm and improvement upon \gls{msgd}, the
de-facto standard for training of \glspl{nn} and variants thereof.  Here,
$\widehat{D}$ refers to the \emph{batch} and $|\widehat{D}| = H \times L$ refers
to the batch size hyper-parameter (e.g. $|\widehat{D}|=64$).  For, $H=1$
\gls{lsgd} degrades to \gls{msgd}.
%
%
The negative effect of data parallelism on the convergence of is a fundamental
property of \gls{msgd}.  An extensive study of this property is presented
by~\citet{shallue2019}.

\paragraph*{\gls{cocoa}~\cite{jaggi2014, smith2018}.}
\label{sec:background:mlprimrt:cocoa}
A state-of-the-art distributed framework for the training of \glspl{glm}.  It is
designed to reduce communication and thus processes significantly more samples
per iteration than, e.g., \gls{msgd}.
We use \gls{cocoa} with a local \gls{scd} solver~\cite{wright15}.  The structure
in \Fig{background:algostructure} is parameterized with $L=1$,
$H=|\widehat{D}|$, whereas $\widehat{D} = D$.  The local update function
$f_{\Delta,k}$ is computed by a local optimizer on partitions $D_{k}$, with $D =
\bigcup_{k=1}^{K} D_{k}$.  In a homogeneous setting each node typically
processes $1/K$-th of the training dataset per iteration.  Data parallelism is
determined by the number of partitions $K$.
Local optimizers detect correlations within the local dataset without global
communication, i.e., the more data is randomly accessible to each optimizer
instance, the less epochs are needed for \gls{cocoa} to converge.  Conversely, if data
access is limited in size, as would be the case when using many tasks, or if no
random access is possible, convergence suffers.
\citet{kaufmann2018} empirically study the relationship between convergence rate
and $K$ and show that by starting with a large $K$ and reducing it after a few
iterations, convergence rate per epoch and time can be increased significantly.


\paragraph*{Summary.}
Both algorithms exhibit an inherent trade-off between data parallelism and
convergence.  Intuitively, a higher degree of parallelism limits the opportunity
to learn correlations across samples, and thus hurts convergence.  While we
focus on two particular methods in this paper, the trade-off between parallelism
and convergence is fundamental in parallel stochastic algorithms.


\subsection{Micro-tasks for distributed training}
%
%

As exemplified in \Fig{introduction:correlation-bs-epochs}, increasing data
parallelism comes at the cost of increasing the total amount of work to
achieve a certain training goal.
Up until a point, the cost increase is smaller than the gain in potential
parallelism, such that overall training time can be reduced by increasing the
data parallelism.  This, however, is only true if and only if all tasks are
executed in parallel.  In shared and heterogeneous environments, this is
generally not true.

Consider the CIFAR-10 example from \Fig{introduction:correlation-bs-epochs}.
For simplicity, we assume perfect linear scaling and zero system overheads.
If one wanted to train on up to 256 nodes, at least 256 tasks are required and
thus a data parallelism of 256 or higher.  According to the data in
\Fig{introduction:correlation-bs-epochs}, this requires 10 epochs to converge.
Assuming that one epoch in this configuration -- where all 256 tasks can run in
parallel -- requires one second, training completes after 10 seconds.
The nature of shared systems is, however, that there are not always enough nodes
available to execute all tasks in parallel.
For instance, let us assume that only 128 nodes are available during the runtime
of the application.  Then each epoch with 256 tasks requires two seconds as two
tasks have to run back to back on each node, resuling in a total training time
of 20s.
If one had used a data parallelism of only 128 from the beginning, instead of
256, training would only require eight epochs or 16s, instead of 20s, resulting
in a training time reduction of 20\%.
This example illustrates the difficulty of elastic scaling of \gls{ml} training
using a micro-task-based system: In many cases, it is only efficient if the
maximal number of nodes (resources) are actually available during most of the
runtime.  This, however, stands in contrast to the goals of elasticity.
This problem is even more pronounced if we also consider load balancing between
differently fast nodes.  The number of tasks required to allow for fine-granular
work redistribution is disproportionately higher than just for elastic scaling
alone.

\section{\Concept for distributed training}
\label{sec:concept}

In the previous section, we showed how micro-tasks inhibit the performance of
distributed training.  In this section we argue that a different execution model, uni-tasks, 
is better suited for \gls{ml} training applications. 
The core idea is very simple: to only use a single task per node.
While this in itself is not a new concept, scientific computing has been 
using MPI that follows this approach for decades, the difficulty is to address the 
scheduling challenges that are typically addressed by micro-tasks, namely elasticity 
and load-balancing. Fortunately, we can exploit the iterative nature of \gls{ml} training 
to tackle these challenges.

%


\paragraph*{Core concepts.}
\Concept consists of two main concepts: immobile tasks and mobile data chunks.
\begin{enumerate}
\item All training samples are stored across a large set of small fixed-sized
  (stateful) data chunks that can be moved between tasks by the scheduler.  Data
  chunks can store dense and sparse training data vectors and matrices of
  variable size.

\item Each node only executes a single task per node (hence the name \concept).
  Each task has full, random access to all training samples across all data
  chunks that are local to a task.
\end{enumerate}

Additionally, a contract between the scheduler and the application is defined
that regulates ownership of a data chunk.
\begin{enumerate}
\item During an iteration, a task owns all task-local data chunks.  It can read
  all and make modifications to data stored in the data chunks, e.g., to update
  per-sample state (e.g., as needed in \gls{cocoa}).  During this period, the
  scheduler does not add or remove data chunks.

\item In-between two iterations, the scheduler owns all data chunks.  Tasks must
  not modify any data chunks and the scheduler is free to add or remove data
  chunks from any task.  Tasks are notified by the scheduler of any data chunk
  addition or removal.
\end{enumerate}

By moving data chunks between tasks in-between iterations, \concept allows one to
add and remove tasks for elastic scaling and to balance load across tasks on
heterogeneous clusters.
\Concept assumes a correlation between the number of training samples in
task-local data chunks and the number of samples processed by each task during
each iteration.
In contrast to micro-tasks, scheduling granularity is determined by the number
of data chunks, not by the number of tasks.  The number of data chunks does not
constitute a lower bound for the level of data parallelism, as multiple data
chunks are processed by the same task, hence the number level of data
parallelism can be lower than the number of data chunks.
In contrast to MPI, \concept defines a method to shift load between tasks.

%

In the following paragraphs, we discuss how elasticity and load balancing are
addressed for distributed training when using \concept.

\paragraph*{Elasticity.}
Elasticity is necessary to efficiently and fairly utilize resources in shared
clusters, to reduce waiting times for job starts, and to react do varying
resource demands of applications throughout their runtime.
We address elasticity in the \concept setting by spawning new tasks as nodes are
added to the application and by terminating them if nodes need to be released.
In both cases data chunks are redistributed across all available tasks.  In the
latter case, however, a prior notification is required such that data chunks can
be transferred before the task is terminated.  Elastic scaling is only possible
in-between iterations.

The application is free to adjust the level of data parallelism during each
iteration to any value equal or larger than the number of tasks.  For both test
applications, we always choose the lowest possible value.

\paragraph*{Load balancing.}
Load balancing is necessary to deal with heterogeneity between cluster nodes as
well as between different hardware (e.g., CPUs vs GPUs) that results in runtime
differences between tasks that process the same amount of input data.

To address heterogeneity, we exploit the fact \gls{ml} training algorithms are
typically iterative and process a known amount of training samples during each
iteration, which allows us to learn how long each task needs to process a
training sample.  \Concept assumes that the number of training samples processed
by each task is a fraction of the total number of training samples across all
task-local data chunks, e.g., a task with twice as many training samples as
another task also processes twice as many per iteration.  This enables the
scheduler to influence the runtime by moving data chunks from tasks on slower to
tasks on faster nodes until their runtime aligns.

As tasks may process a different number of training samples during each
iteration, their model updates need to be weighted differently as well (as
proposed in \citet{stich2018}).  We do this by multiplying the model update
$f_{\Delta,k}$ of task $k$ by $D_k/\widehat{D}$ (see
\Eq{background:updatecomputation}).

\section{\Sys design and implementation}
\label{sec:system}
Here, we describe how \Sys implements an elastic distributed training framework
using \concept.

\begin{figure}[t!hbp]
  \centering \includegraphics[width=0.8\linewidth]{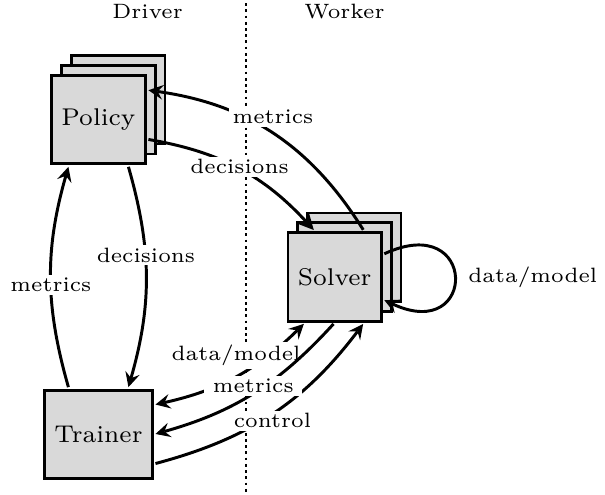}
  \caption{High-level architecture of \Sys.}
  \label{fig:system:overview}
\end{figure}

\subsection{Overview}
\label{sec:system:overview}

\Sys, as shown in \Fig{system:overview}, is based on a driver/worker design with
a central driver (\emph{trainer}) and multiple workers (\emph{solvers})
communicating via a RDMA-based RPC mechanism (see \Sec{system:commsubsys}).
The driver executes the \emph{trainer} module, which, in tandem with multiple
policy modules, is responsible for coordinating training.  Policy modules make
scheduling decisions, such as assigning chunks, balancing load, and scaling in
and out.
Worker processes execute \emph{solver} modules (\concept) and implement the
\gls{ml} algorithms (e.g., \gls{scd} for \gls{cocoa}).  Crucially, only a single
(multi-threaded) worker process is executed per node.  Solvers are controlled by
the trainer and policy modules, which in turn receive model and state updates as
well as metrics (e.g., duality-gap).

\Sys applications need to implement a trainer and solver module, and may
optionally implement policy modules to control system behavior during training.
For instance, our \gls{lsgd} implementation uses libtorch (from
PyTorch~\cite{paszke2017}) in the solver for forward and backward propagation
steps.  The trainer module acts as synchronous parameter server that merges
updates from solver instances.  A simplified version of the \gls{lsgd} code is
shown in \Lst{system:app}.

In the remainder of this section, we elaborate on each module as well as
the communication subsystem and in-memory data format of \Sys.

\subsection{Trainer and solver}
\label{sec:system:trainer_solver}
The trainer and solver modules represent application code.  Trainer modules are
the central controlling entity and coordinate individual solver instances in
tandem with policy modules.  Policy modules can implement complex (reusable)
optimizations (e.g., online hyper-parameter tuning), and solver modules
implement arbitrary functions for distributed execution.  Only a single solver
module is executed per node and application, therefore, each solver module can
internally spawn threads and use all CPUs or GPUs of a node.
Trainer and solver modules periodically synchronize at global barriers,
e.g. in-between iterations, but can exchange additional messages at any time.

\subsection{Communication subsystem}
\label{sec:system:commsubsys}
In distributed training, communication can easily become a bottleneck. For
example, using \gls{cocoa} to train a model for the Criteo dataset (see
\Tab{evaluation:datasets}), each task has to send/receive $\approx$16MiB in
updates in-between iterations.
For that reason, we built our communication subsystem on RDMA.  RDMA allows
low-overhead, zero-copy, one-sided operations for bulk data transfers, such as
model and training (input) data as well as two-sided \glspl{rpc} using RDMA
send/receive.


\subsection{In-memory chunk data format}
\label{sec:system:dataformat}
To fully exploit RDMA, data is stored in static, consecutive memory regions.
The in-memory representation of training (input) data is based on fixed-sized
data chunks.  Chunks can store sparse or dense training data vectors and
matrices.  The number of training samples per data chunk can vary depending on
their size.  Chunks allow to easily move training data subsets between nodes.
The chunk size can be tuned to an optimal value depending on dataset and system
properties, e.g. to the CPU cache size.

\Sys's in-memory format is application agnostic and simply provides applications
a contiguous memory space that can be moved across nodes in-between iterations.
For instance, our \gls{lsgd} implementation stores the backing memory of native
PyTorch tensor objects in data chunks wheras for \gls{cocoa}, we simply store
sparse vectors as well as per-sample state in a data chunk.  Having the ability
to store per-sample state in a data chunk is important as it ensures that state
and the data it correlates to are always moved together.

One important limitation of \Sys's data chunk is that they must not require any
serialization, as one-sided RDMA read operations are used to transfer them.
Deserialization is possible.  In the case of PyTorch, for instance, we restore
tensor objects via the \texttt{torch::from\_blob} function, which creates a new
tensor object backed by the in-chunk data.


%
%


\subsection{Policies}
\label{sec:system:policies}
\Sys implements a flexible policy framework which we use to implement vital
parts of the system.  Policies make decisions based on events and metrics they
receive from trainer and solver modules and return proper decisions for them.
Each policy module runs in a separate thread and multiple policy modules can run
at the same point in time.  Policy modules coordinate with the trainer and can
coordinate with each other. Next, we present the most relevant policy modules.

\paragraph{Elastic scaling policy.}
\label{sec:system:policies:scaling}
This module interfaces with the resource manager, e.g.,
YARN~\cite{vavilapalli2013}, to make resource requests and get resource
assignment and revocation notices.
Upon receiving a new resource assignment, it registers a new worker
(task) and notifies the trainer.  After the current iteration, it shifts data
chunks from old to new workers.  It relies on the rebalancing policy
(\Sec{system:policies:rebalancing}) to ensure proper load balancing.
\Sys expects the resource manager to give advance notice before revoking a
resource allocation.  Upon receiving such a notice, it redistributes data chunks
from to-be freed workers to remaining ones in a round robin fashion.  As before,
it relies on the rebalancing policy to ensure load balance.

\paragraph{Rebalancing policy.}
\label{sec:system:policies:rebalancing}
The rebalance policy observes iteration runtimes over multiple iterations to
learn the per-sample runtime of each task, as described above.
Between iterations, solvers are ranked according to their median performance
over the last $I$ iterations and chunks moved gradually, across multiple
iterations, from slower to faster solvers until performance differences are
smaller than the estimated processing time of a single chunk.
This policy can also be used to address slowly changing performance of nodes,
e.g. ones that are caused by the start/end of long running background jobs and
restore balance after scaleing in and out.  Its robustness against runtime
fluctuations can be adjusted by tweaking $I$.

We decided against reloading data from a (shared) filesystem as data loading
turned out to be more expensive than transferring loaded data between nodes,
especially if input files are stored on a shared network filesystem.  Moreover,
our in-memory format can combine data chunks with the corresponding state, which
needs to be transferred between workers anyway.

\paragraph{Other policies.}
\label{sec:system:policies:other}
Apart from the above described policies, we have implemented policies for
straggler mitigation, global background data shuffling and others.

\section{Evaluation}
\label{sec:evaluation}
Our evaluation shows how \sys performns in an elastic setting where nodes are
added and removed during training and on a heterogeneous cluster where nodes are
differently fast.
As no other elastic, load-balancing \gls{ml} training framework is publicly
available, we emulate micro-tasks with \sys.
Additionally, we compare \sys with two state-of-the-art rigid \gls{ml} training
frameworks in non-elastic, non-heterogeneous scenario to establish a performance
baseline.

%




\subsection{Evaluation setup and methodology}
\label{sec:eval:setup}
\label{sec:eval:setup:cluster}
Our test cluster consists of 16+1 nodes.  Nodes are equipped with Intel Xeon
E5-2630/40/50 v2/3 with 2.4 -- 2.6GHz and 160 -- 256GiB RAM.  We execute \Sys
inside Docker containers.  For some heterogeneity experiments, we reduce the CPU
frequency of four nodes from 2.6 to 1.2GHz.
All nodes are connected by a 56GBit/s Infiniband network via a Mellanox SX6036
switch.  During experiments, up to 16 nodes are used for workers and one node
for the \Sys driver.

Our test applications are \gls{lsgd} and \gls{cocoa}, using test accuracy as a
metric for convergence for the former and the duality-gap~\cite{jaggi2014,
  smith2018} for the latter.  We train on each dataset for $\approx$20 minutes,
after which we terminate the training.  Each experiment is repeated five times
and average results are presented.
\Tab{evaluation:datasets} lists all datasets we use during the evaluation.  The
chunk size is set to 1MiB in \gls{cocoa} experiments and, due to the smaller
dataset sizes, 200KiB for \gls{lsgd} experiments.
\label{sec:evaluation:datasets}
\begin{table}[h!]
  \caption{Number of samples (\#S), features (\#F) and categories (\#C) of
    datasets used in the evaluation. Size is given for the in-memory
    representation.}
  \label{tab:evaluation:datasets}
  \vskip 0.15in
  \begin{center}
    \begin{small}
      \begin{sc}
        \begin{tabular}{l|rrrr}
          \toprule
          Dataset       &  \#S &  \#F & \#C &   Size \\
          \midrule
          Higgs         &  11M &   28 &   2 & 2.5GiB \\
          Criteo        &  46M &   1M &   2 &  15GiB \\
          CIFAR-10      &  60k & 3072 &  10 & 162MiB \\
          Fashion-MNIST &  70k &  784 &  10 &  30MiB \\
          \bottomrule
        \end{tabular}
      \end{sc}
    \end{small}
  \end{center}
  \vskip -0.1in
\end{table}

\paragraph*{Synchronous local \gls{sgd}.}
We implemented \gls{lsgd}~\cite{lin2018} for \Sys based on libtorch, the C++
backend of PyTorch~\cite{paszke2017}.
We train a \gls{cnn} with relu activation composing of two convolutional layers
with max-pooling followed by 3 fully connected layers on the CIFAR-10 and
Fashion-MNIST datasets using \gls{lsgd}.  We use $L=8$ and $H=16$, a momentum of
0.9 and a base learning rate $\alpha$ of 1e-4 for CIFAR-10 and 5e-4 for
Fashion-MNIST.  According to best practice, we scale the learning rate with the
square root of the number of tasks $K$ such that the effective learning rate
$\alpha' = \alpha \times \sqrt{K}$.
The global batch size (number of samples processed during each iteration across
all tasks) is $K \times L \times H$.  For micro-tasks, we select four values for
$K=\{16,24,32,64\}$.  Using different values of $K$ allows us to assess the
trade-off between scheduling and algorithmic efficiency.  Here, K remains
constant during the training.
For \concept $K$ equals number of currently used nodes.
As \gls{msgd} is a special case of \gls{lsgd} with $H=1$ we trivially also
support \gls{msgd}, which we use for baseline comparisons with PyTorch.

\paragraph*{\glstext{cocoa}.}
We implemented \gls{cocoa} with a local \gls{scd} solver for \Sys based on the
original Spark implementation~\cite{cocoa_source}.
We train a \gls{svm} on the Higgs and Criteo datasets.  We use \gls{scd} as
local solver with $L=1$ and $H$ equal to the number of local training samples.
The number of tasks $K$ is the same as above.  The algorithm parameter $\sigma$
is set to the the number of tasks, and the regularization coefficient $\lambda$
to the number of samples $\times$ 0.01.

\paragraph*{Micro-tasks.}
As no elastic \gls{ml} training framework based on micro-tasks (or any other
technique) is publicly available and general-purpose frameworks such as Spark do
not perform competitively~\cite{dunner2017}, we emulate micro-tasks using \Sys
with a constant number of tasks $K$ and measure the convergence rate per epoch.
It is possible to do this accurately because in micro-tasks, convergence rate
per epoch only depends on the number of tasks but not on the number of nodes or
on which node a task is executed on.  It does not, however, allow us to directly
measure the convergence rate over time for micro-tasks.  Instead, we project the
latter by assuming an optimal schedule for the number of tasks, nodes and
relative node performance.  Henceforth, the number of micro-tasks is given in
parentheses.


Using \Sys to emulate micro-tasks during elasticity and load balancing
experiments has the additional benefit of keeping implementation-specific
variables, such as the implementation of the training algorithms (\gls{lsgd} and
\gls{cocoa}), the communication subsystem (e.g., RDMA vs. TCP/IP), and other
factors constant.

\subsection{Baseline comparisons}
\label{sec:evaluation:reference}
We compare \sys against Snap~ML~\cite{dunner2018_2} for \gls{cocoa} and
PyTorch~\cite{paszke2017} for \gls{msgd} in a non-elastic, non-heterogeneous
scenario using the same training algorithms, hyper-parameter values and datasets
on the test setup described above.  None of the novel functionality of \Sys was
used in this experiment.  The purpose of this experiment is to show that \Sys
does not impair performance in the \emph{normal} non-elastic, non-heterogeneous
case.
%
We measure convergence rate per epoch and over time.  Detailed results of this
experiment are provided in \Sec{evaluation:reference} and are summarized here.

Convergence behavior per epoch for \gls{msgd} is identical on \Sys and PyTorch
while \Sys requires slightly less time per epoch.
Compared to Snap~ML, \Sys performed virtually identically for the Higgs dataset
but outperformed it for the Criteo dataset due to differences in data
partitioning.
This experiment confirms that \Sys's baseline performance is on par with that of
highly optimized, established \gls{ml} training frameworks.  In contrast to
those, \Sys is able to elastically scale during execution and balance load in
heterogeneous clusters.  Both aspects are evaluated in the following.

\subsection{Elastic scaling}
\label{sec:evaluation:scaling}
In this section, we evaluate \sys with the elastic scaling policy enabled in two
elastic scenarios and compare it to micro-tasks.
Specifically, we consider:
\begin{enumerate*}[label=\bfseries\roman*)]
\item the effect of \emph{data parallelism} (batch size for \gls{lsgd}, and
  number of partitions for \gls{cocoa}) on the number of epochs to converge,
  and
\item the trade-off between scheduling efficiency and convergence under
  micro-tasks.
\end{enumerate*}

\paragraph*{Methodology.}
Our test scenarios consist of gradual scale-in from 16 to 2 nodes and scale-out
from 2 to 16 nodes.  We add (remove) 2 nodes every 20s until the maximum
(minimum) number of nodes is reached.  During each run, we measure convergence
per epoch and project convergence over time using an optimal schedule for
\concept and micro-tasks for each number of nodes.
In micro-tasks, elastic scaling works by distributing a fixed number of tasks
across more or fewer nodes and not by adjusting the number of tasks.  Moreover,
the number of nodes is typically not known by the application.  Hence, we assume
a fixed number of tasks independently of the nodes used.  To project the time
per iteration, we assume a normalized task runtime (one task, processing
$1/16$th of the data takes one time unit) and compute the number of task waves
necessary for each iteration.
\begin{itemize}
\item $K$ micro-tasks on $N$ nodes require $\lceil K/N \rceil$ task waves, as
  only $N$ tasks can be executed at the same time.  In consequence, each
  iteration requires $16/K \times \lceil K/N \rceil$ time units.  For instance,
  $K=32$ tasks on $N=14$ nodes require $\lceil 32/14 \rceil = 3$ task waves and
  $16/32 \times 3 = 1.5$ time units per iteration.

\item For \gls{cocoa} on \concept, load is redistributed such that a single
  iteration takes $16/N$ time units.  For instance, on 14 nodes, one iteration
  requires $16/14=1.14$ time units.  For \gls{lsgd} on \concept, the batch size
  is adjusted such that each iteration still only requires one time unit.
  Instead, the number of iterations per epoch increases by $16/N$.
\end{itemize}

Our time projections do not include data transfer overheads.  As each task needs
to communicate model updates, the total communication volume of micro-tasks is
as least as high as that of \concept, hence by ignoring data transfer overheads,
we favor micro-tasks.

%
\paragraph*{Results.}
\begin{figure*}[t]
  \centering
  \subfloat[CIFAR-10 (\glstext{lsgd})] {%
    \parbox{0.245\linewidth}{
      \includegraphics[width=\linewidth]{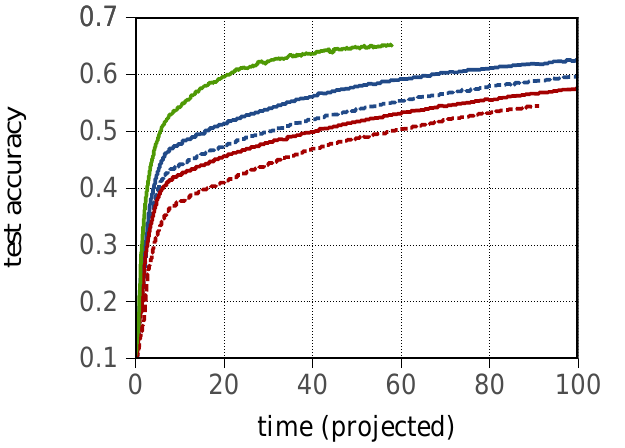}

      \includegraphics[width=\linewidth]{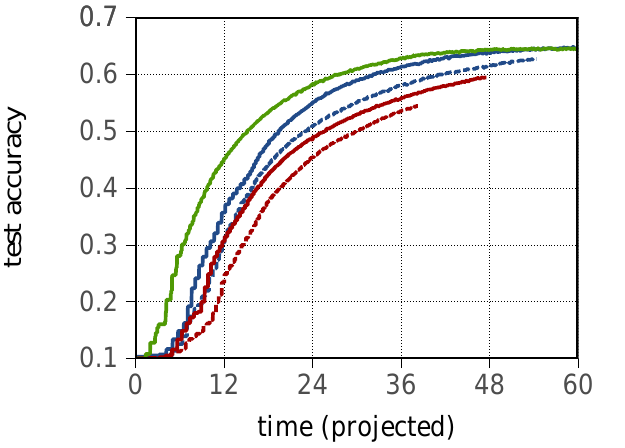}
      \vspace{-0.2cm}
      \label{fig:evaluation:scaling:convergence-per-time:cifar-10}
    }
  }
  \subfloat[Fashion-MNIST (\glstext{lsgd})] {%
    \parbox{0.245\linewidth}{
      \includegraphics[width=\linewidth]{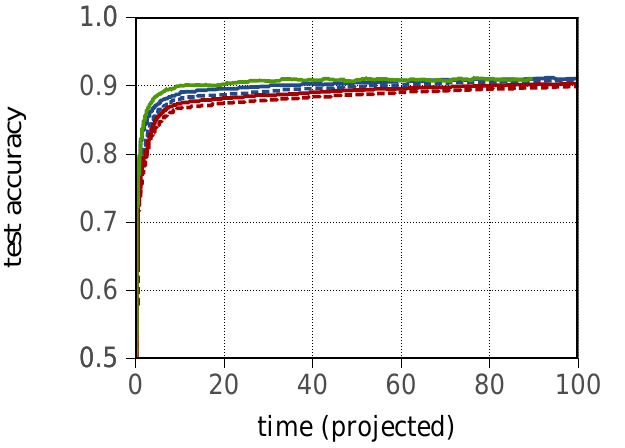}

      \includegraphics[width=\linewidth]{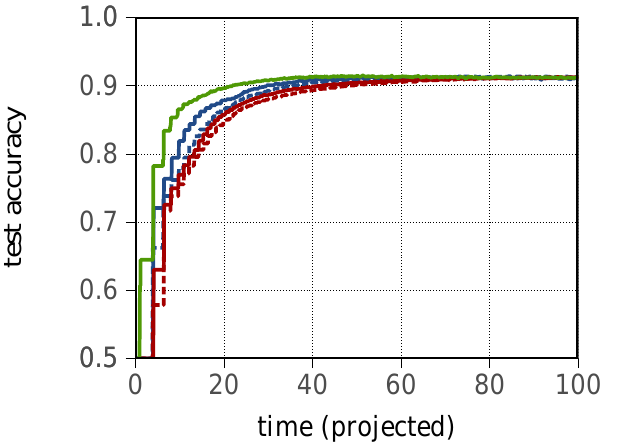}
      \vspace{-0.2cm}
      \label{fig:evaluation:scaling:convergence-per-time:fashion-mnist}
    }
  }
  \subfloat[Higgs (\glstext{cocoa})] {%
    \parbox{0.245\linewidth}{
      \includegraphics[width=\linewidth]{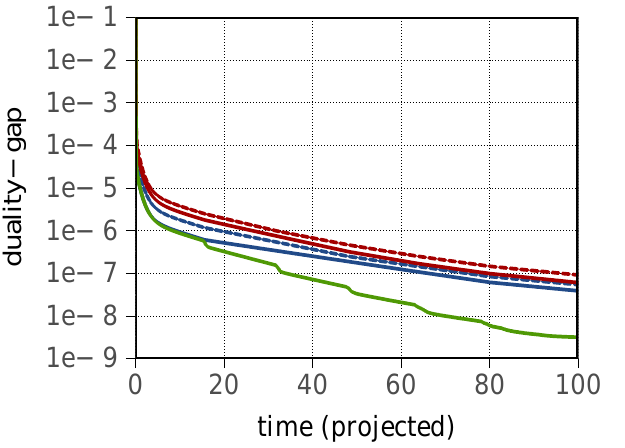}

      \includegraphics[width=\linewidth]{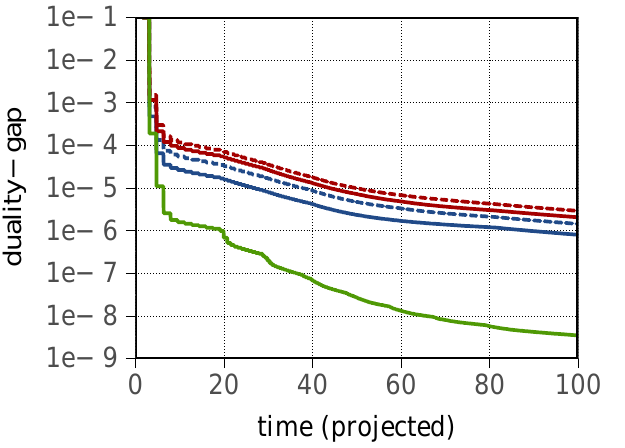}
      \vspace{-0.2cm}
      \label{fig:evaluation:scaling:convergence-per-time:higgs}
    }
  }
  \subfloat[Criteo (\glstext{cocoa})] {%
    \parbox{0.245\linewidth}{
      \includegraphics[width=\linewidth]{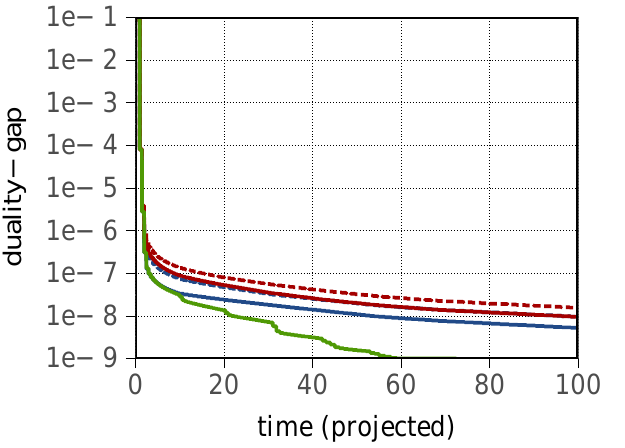}

      \includegraphics[width=\linewidth]{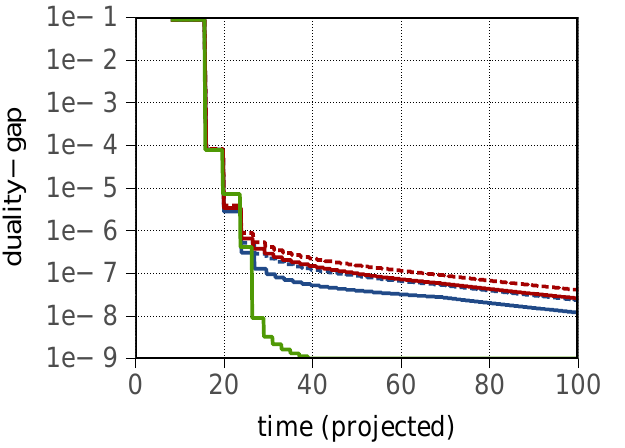}
      \vspace{-0.2cm}
      \label{fig:evaluation:scaling:convergence-per-time:criteo}
    }
  }
  \vspace{0.2cm}
  \includegraphics[scale=1.0]{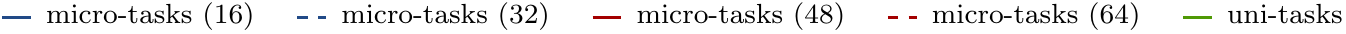}
  \caption{Convergence over time (projected) vs. data parallelism for elastic
    \textbf{scale-in (top)} and \textbf{scale-out (bottom)} experiments.  The
    number of micro-tasks is given in parentheses.  Time is normalized to 100
    time units.}
  \label{fig:evaluation:scaling:convergence-per-time}
\end{figure*}

\Fig{evaluation:scaling:convergence-per-time} shows detailed convergenve over
time plots for elastic scale-in and out for different data parallelism values.
Convergence per epoch results are provided in the appendix
(\Sec{appendix:evaluation:scaling}).
Generally, the higher the data parallelism, the more epochs are needed to
converge for micro-tasks, which is consistent with our initial problem statement
and previous studies~\cite{shallue2019}.
As \Fig{evaluation:scaling:convergence-per-time} shows, the increased scheduling
efficiency of using more micro-tasks cannot compensate for the reduced
convergence rate per epoch and micro-tasks (16) consistently outperforms other
micro-tasks configurations.

Moreover, the convergence rate over time with \concept is equal or higher during
scale-in and -out, showing that the ability to adjust the level of data
parallelism across a wide range can improve convergence per epoch and over
time.\footnote{Applications are free to choose any level of data parallelism
  equal or larger the number of tasks if it benefits convergence.}  This ability
is not only beneficial in shared environments but can also be exploited to
accelerate the training process in general.  \citet{kaufmann2018} show for
\gls{cocoa}, that scaling in training at specific points in time can accelerate
training by up to 6$\times$.  \citet{smith2017_2} report that increasing the
batch size as alternative to reducing the learning rate once convergence slows
down is beneficial for mini-batch \gls{sgd}.  Both cases could be implemented
with \sys.


However, results differ across algorithms and datasets.
For \gls{lsgd}, scale-in as well as scale-out on \concept improves convergence
over time compared to the best micro-tasks configuration.  In the scale-out
case, the global batch size for \concept is smaller in the beginning but
equalizes with micro-tasks (16) quickly as nodes are added.  In the scale-in
case, the global batch size for \concept is the same as for micro-tasks (16) in
the beginning but is quickly reduced.  As it is smaller for longer, compared to
the scale-out case, the convergence benefits over micro-tasks (16) are higher in
the scale-in case.

%
%
%
%
%

The average maximal test accuracy for \concept is virtually identical to that of
micro-tasks (16), which is the best micro-tasks configuration in all but one
case: In the scale-in case for CIFAR-10, \concept achieves an average maximal
test accuracy of 65.6\% compared to 65.0\% for micro-tasks (16).

%
Results for \gls{cocoa} are similar.  
Scaling in reduces the number of epochs as well as time to converge, as
suggested in \citet{kaufmann2018}.  After each scale-in step (which can be
identified in \Fig{evaluation:scaling:convergence-per-time:higgs} and
\Fig{evaluation:scaling:convergence-per-time:criteo}) convergence rate improves.
The reason for this behavior is that the local \gls{scd} solver has access to
additional training data and can therefore identify new correlations across
training samples locally.
Scaling out behaves similarly which is, at first sight, counter intuituve as
every task gets to see fewer and fewer training samples as training scales out.
However, during scale-out the data chunks that are moved to newly added tasks
are picked randomly from each old task which effectively shuffles training
samples.  This also allows the solver to identify new correlations locally while
also decreasing the duration of each iteration.

\subsection{Load balancing}
\label{sec:evaluation:loadbalancing}
\begin{figure*}[t]
  \centering
  \subfloat[CIFAR-10 (\glstext{lsgd})] {%
    \parbox{0.245\linewidth}{
      \includegraphics[width=\linewidth]{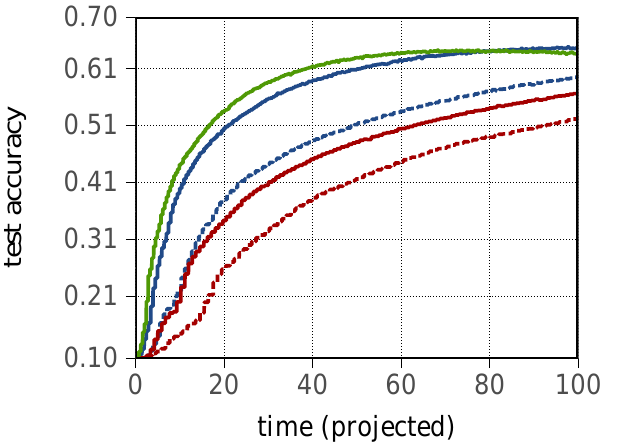}
      \vspace{-0.2cm}
      \label{fig:evaluation:balance:convergence-per-time:cifar-10}
    }
  }
  \subfloat[Fashion-MNIST (\glstext{lsgd})] {%
    \parbox{0.245\linewidth}{
      \includegraphics[width=\linewidth]{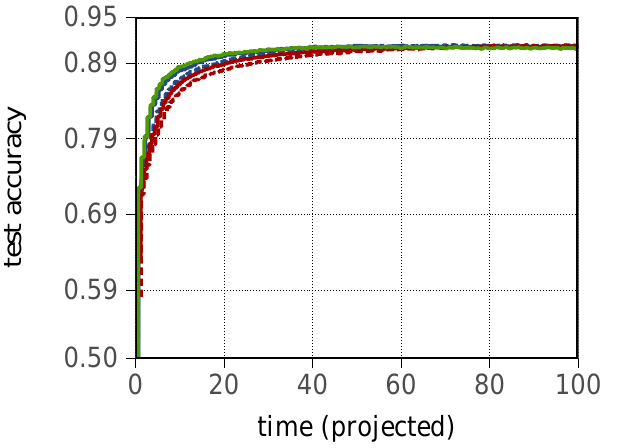}
      \vspace{-0.2cm}
      \label{fig:evaluation:balance:convergence-per-time:fashion-mnist}
    }
  }
  \subfloat[Higgs (\glstext{cocoa})] {%
    \parbox{0.245\linewidth}{
      \includegraphics[width=\linewidth]{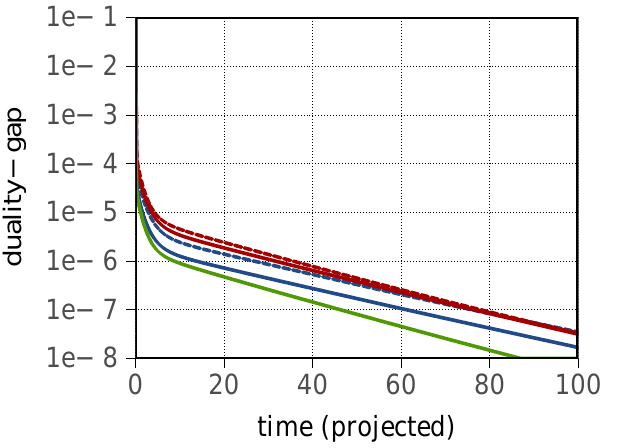}
      \vspace{-0.2cm}
      \label{fig:evaluation:balance:convergence-per-time:higgs}
    }
  }
  \subfloat[Criteo (\glstext{cocoa})] {%
    \parbox{0.245\linewidth}{
      \includegraphics[width=\linewidth]{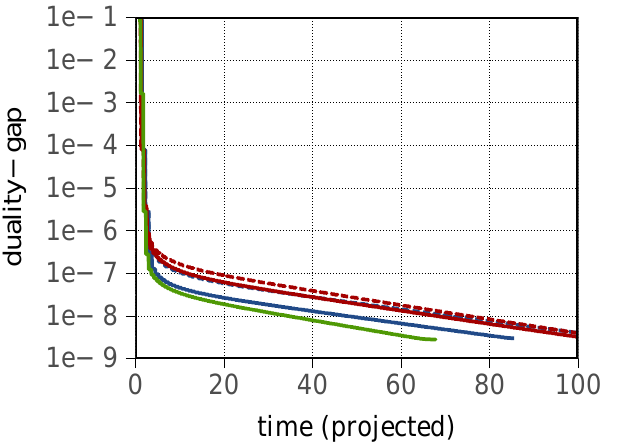}
      \vspace{-0.2cm}
      \label{fig:evaluation:balance:convergence-per-time:criteo}
    }
  }
  \vspace{0.2cm}
  \includegraphics[scale=1.0]{figures/legend}
  \caption{Convergence over time (projected) when balancing load balancing in a
    heterogeneous cluster.  The number of micro-tasks is given in parentheses.
    Time is normalized to 100 time units.}
  \label{fig:evaluation:balance:convergence-per-time}
\end{figure*}
In this section, we compare \Sys, with the load balancing policy enabled, to
micro-tasks in a heterogeneous scenario with nodes of different speed.  Such a
scenario can occur in practice, as compute clusters are often not replaced
completely but extended and partially replaced over time using multiple
generations of hardware (e.g., CPUs, GPUs)~\cite{delimitrou2014}.  Even the same
cloud instance type can be backed by different models and generations of
hardware~\cite{ou2012}.

In a heterogeneous scenario, faster nodes should perform more of the overall
work than slower nodes, such that all nodes finish at the same time for each
iteration.  In a micro-task based system, this is achieved by scheduling more
tasks on fast nodes than slow nodes.
This, however, requires multiple tasks to be executed per node so that one or
more of them can be moved to other nodes.  In consequence, no load balancing is
possible with micro-tasks (16) on our 16 node test cluster.
%
%
\Sys balances load by shifting data chunks, of which there are typically
hundreds or thousands, from slow nodes to fast nodes and by adjusting the number
of samples that individual \concept process in each iteration, such that all
tasks finish at the same time, independently of the node performance.

\paragraph*{Methodology.}
We evaluate heterogeneous load balancing in two scenarios:
\begin{enumerate*}[label={\arabic*)}]
\item We configure the load balancing policy of \Sys to assume eight fast and
  eight slow nodes, with the latter being 1.5$\times$ slower than the former and
  measure the number of epochs to converge.  This simple scenario allows us to
  project time to convergence.
  
\item We execute \Sys with the load balancing policy enabled on our test cluster
  where the CPU frequency of four nodes has been reduced to increase the level
  of heterogeneity.  We measure the task and iteration runtimes as well as the
  number of data chunks of each task across the load balancing process to show
  how \Sys can correctly learn task runtime and balance load in response.
\end{enumerate*}


%
In micro-tasks, load balancing works by balancing fixes-size tasks across all
nodes and not by adjusting the number training samples per task.  Hence, we
assume that each task processes the same number of training samples per
iteration.  To project the time per iteration, we assume a normalized task
runtime: One task, processing $1/16$th of the data takes one time unit on the
fast nodes and 1.5 time units on the slow nodes.  We use this to compute the
optimal (shortest) schedule for each iteration.
\begin{itemize}
\item For micro-tasks, $K$ tasks on eight fast and eight slow nodes, the optimal
  schedule is $\max(i \times 1.5s, j \times 1.0s) \times 16/K$ long with $i$
  ($j$) being the number of tasks on each slow (fast) node such that the
  schedule length is minimal.  For instance, with $K=64$ tasks, the optimal
  schedule is $\max(3 \times 1.5s, 5 \times 1.0s) \times 16/64 = 1.25s$ per
  iteration.

\item For \concept, load is redistributed such that fast nodes process
  1.5$\times$ as many training samples as slow nodes, resulting in an iteration
  duration of $1.2s$.
\end{itemize}

As before, our time projections do not include data transfer overheads, which
favors micro-tasks.

\paragraph*{Results.}

\Fig{evaluation:balance:convergence-per-time} shows detailed convergenve over
time plots for different data parallelism values.
Convergence per epoch results are shown in \Sec{appendix:evaluation:balance}.
Per epoch, \Sys converges as fast as micro-tasks (16).  Over time, however, \Sys
converges faster than any micro-tasks configuration as it requires as few epochs
to converge as micro-tasks (16) but can balance load more effectively than
micro-tasks (64), which reduces iteration duration and thus combines algorithmic
and scheduling efficiency.  For \gls{lsgd}, the average maximal test accuracy is
$\approx$0.5\% lower with \concept than with micro-tasks (16).  However, no load
balancing is actually possible with the latter.  Compared to other micro-task
configurations, \concept achieves a similar average maximal test accuracies.
For \gls{cocoa}, \concept converges virtually identical to micro-tasks (16) per
epoch but outperforms it over time due to its ability to balance load more
effectively.

\begin{figure}[t]
  \centering
  \includegraphics[width=0.7\linewidth]{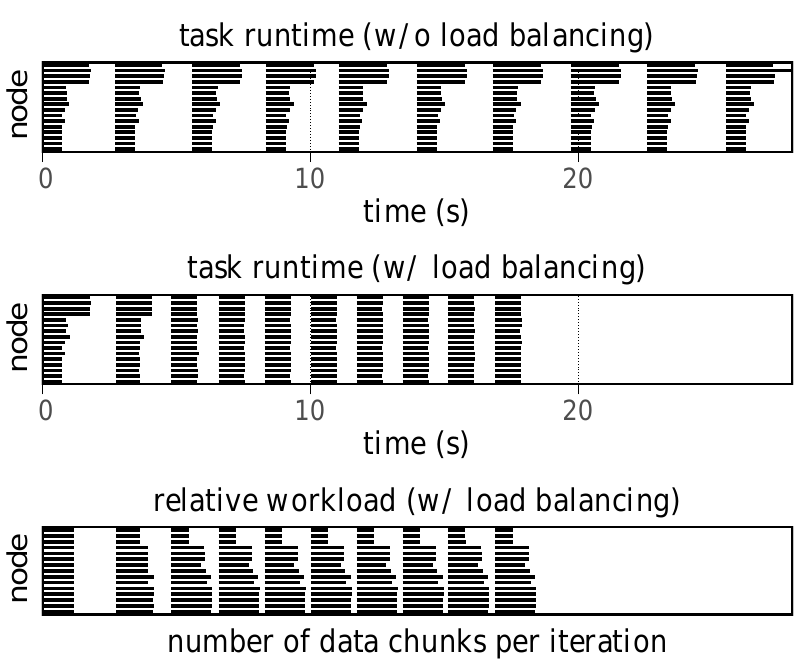}
  \caption{Visualization of the load balancing process on a real heterogeneous cluster.}
  \label{fig:evaluation:balance:process}
\end{figure}

Swimlane diagrams in \Fig{evaluation:balance:process} visualize the load
balancing process for the Criteo dataset on our test cluster where the CPU
frequency of four nodes has been reduced to 1.2GHz to improve the visibility of
this process.  Results for the other datasets are similar and provided in the
appendix (\Sec{appendix:evaluation:balance}).

The top diagram shows task runtimes per node and iteration without load
balancing.  Here, iteration duration is determined by the four slow nodes.  Task
runtimes are visualized by horizontal black bars.  Bars that start at the same
time represent tasks of the same iteration.  Space in-between bars represents
time during which tasks are inactive, i.e., communicating or waiting for the
latest model update from the trainer.
The middle diagram shows task runtimes with load balancing enabled.  During the
first iteration, task runtimes are the same as without load balancing.  As load
is shifted during subsequent iterations, task runtimes align and iteration
durations reduce.
The bottom diagram shows the relative workload (not time) of tasks in the middle
diagram.  It shows how the workload is shifted from slow to fast nodes.  The
length of the bars represent the number of data chunks for each task and
iteration, relative to all other tasks and iterations.
After a few iterations, workload and task runtimes stabilize as \Sys has learned
the performance of each node and balance load accordingly.

\section{Conclusion and future work}
\label{sec:conclusion}

We presented \Sys, a distributed \gls{ml} training framework based on \concept.
\Sys enables efficient elastic scaling and load balancing without incurring
overheads that are typical for micro-task systems and can thereby accelerate
time to convergence by orders of magnitude in some cases.
Our work touches many issues that distinguish distributed \gls{ml} training from
regular distributed applications, such as their sensitivity to data parallelism.
Still, many aspects of \gls{ml} workloads remain unexplored, and we believe
there is a lot of potential to further exploit the unique properties of \gls{ml}
algorithms to build more efficient systems.


%
%



\bibliography{bibliography}
\bibliographystyle{sysml2019}

\clearpage
\appendix
\section{Appendix}
\label{sec:appendix}

\subsection{Baseline comparisons}
\label{sec:evaluation:reference}
We compare \sys against Snap~ML~\cite{dunner2018_2} for \gls{cocoa} and
PyTorch~\cite{paszke2017} for \gls{msgd} in a non-elastic, non-heterogeneous
scenario.
Neither compared-to framework is able to elastically scale nor balance load.
The purpose of this comparison is to show that the elasticity and load balancing
capabilities of \sys and \concept do not come at a cost of performance in the
\emph{normal} case.  In consequence, \Sys's elasticity and load balancing
policies are also not used during these experiments.
Both frameworks are executed with RDMA-enabled MPI communication backends.
We measure the convergence per epoch and over time.  Each experiment is repeated
5$\times$.

\paragraph*{\ok{PyTorch}.}
As no \gls{lsgd} implementation for PyTorch exists, we compared \Sys to PyTorch
using \gls{msgd}.  \gls{msgd} is a special case of \gls{lsgd} with $H=1$.
\Sys's \gls{msgd} training algorithm uses libtorch, the C++ backend of PyTorch,
which allows us to rule out the implementation of the training algorithm as
source for any potential differences.
For both datasets, a learning rate of 0.002 and a momentum of 0.9 is used.

Convergence per epochs is virtually identical to PyTorch.  This is expected as
both are based on libtorch and therefore use the same training algorithm
implementations, \gls{cnn} and hyper-parameters.  Per time, \Sys is slightly
faster, which is likely due to overheads introduced by Python, which do not
afflict \sys, at it is natively implemented in C++.  The maximal test accuracy
that was achieved within the test duration is \ok{65.2\%} for CIFAR-10 with
both frameworks.  For Fashion-MNIST, \Sys has a \ok{0.2\%} lead over PyTorch
with \ok{91.4\%}.
Note that we did not tune hyper-parameters for each dataset dataset nor adjust
them online, which is why the test accuracy for CIFAR-10 degrades slightly after
reaching a peak.

\paragraph*{\ok{Snap~ML}.}
\Sys's \gls{cocoa}/\gls{scd} implementation for the training of a \gls{svm} is
based on the original Spark implementation~\cite{cocoa_source}.  The algorithm
parameter $\sigma$ is set to the the number of tasks, and the regularization
coefficient $\lambda$ to the number of samples $\times$ 0.01.
Compared to Snap~ML, \Sys shows similar convergence and runtime behavior for the
Higgs dataset.  For Criteo, however, \Sys converges much faster.  This is due to
the sensitivity of Criteo to data partitioning.  \Sys randomly assigns data
chunks to tasks, whereas Snap~ML splits the data into 16 contiguous partitions.
Per iteration, Snap~ML is slightly faster as \Sys's reduce and broadcast
primitives are less optimized than their MPI counterparts used by Snap~ML.

With one exception (Criteo), \Sys performs similarly to both rigid frameworks in
a baseline scenario, showing that neither \Sys not \concept impair baseline
performance.

\begin{figure}[H]
  \centering \subfloat[CIFAR-10 (epochs)] {%
    \includegraphics[width=0.49\linewidth]{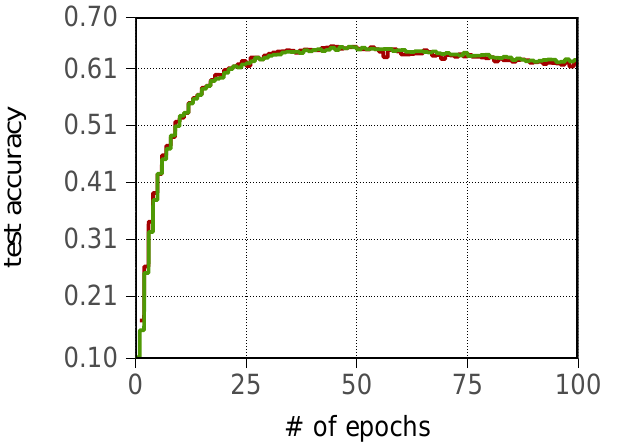}
    \label{fig:evaluation:reference:convergence-per-epoch:cifar-10}
  }
  \subfloat[Fashion-MNIST (epochs)] {%
    \includegraphics[width=0.49\linewidth]{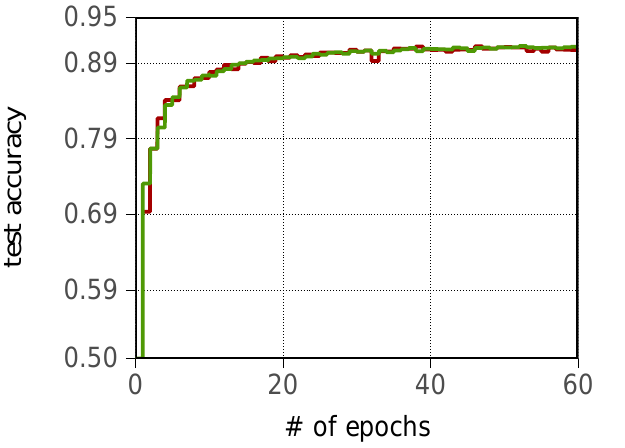}
    \label{fig:evaluation:reference:convergence-per-epoch:fashion-mnist}
  }

  \centering \subfloat[CIFAR-10 (time)] {%
    \includegraphics[width=0.49\linewidth]{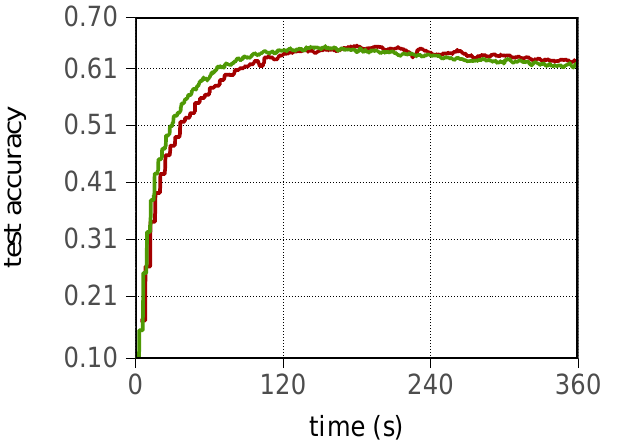}
    \label{fig:evaluation:reference:convergence-per-time:cifar-10}
  }
  \subfloat[Fashion-MNIST (time)] {%
    \includegraphics[width=0.49\linewidth]{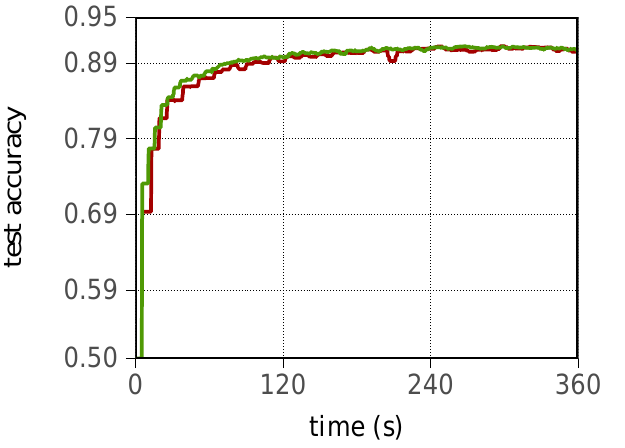}
    \label{fig:evaluation:reference:convergence-per-time:fashion-mnist}
  }
  \caption{\ok{Comparison with PyTorch w.r.t.\ convergence over epochs (top) and time (bottom).}}
  \label{fig:evaluation:reference:convergence-per-epoch}
  \label{fig:evaluation:reference:convergence-per-time}
\end{figure}

\begin{figure}[H]
  \subfloat[Higgs (epochs)] {%
    \includegraphics[width=0.49\linewidth]{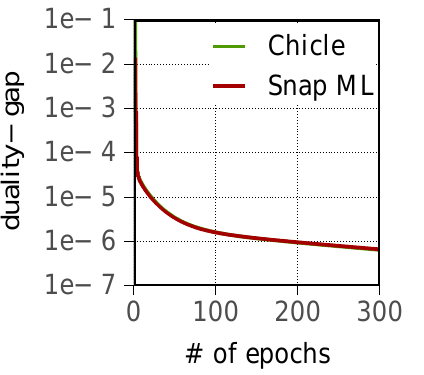}
    \label{fig:evaluation:reference:convergence-per-epoch:higgs}
  }
  \subfloat[Criteo (epochs)] {%
    \includegraphics[width=0.49\linewidth]{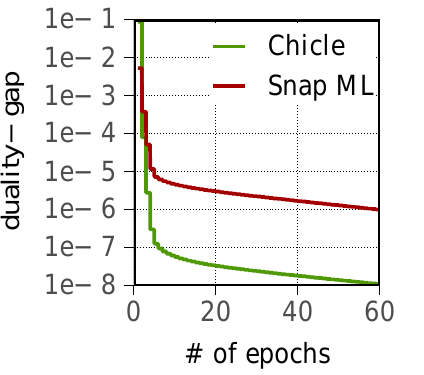}
    \label{fig:evaluation:reference:convergence-per-epoch:criteo}
  }

  \subfloat[Higgs (time)] {%
    \includegraphics[width=0.49\linewidth]{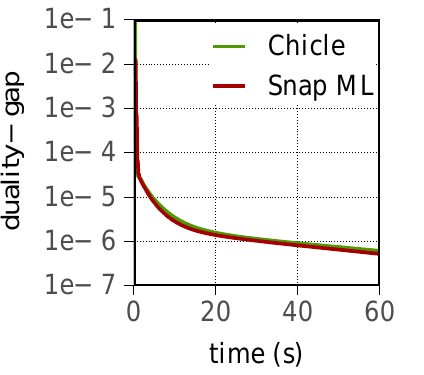}
    \label{fig:evaluation:reference:convergence-per-time:higgs}
  }
  \subfloat[Criteo (time)] {%
    \includegraphics[width=0.49\linewidth]{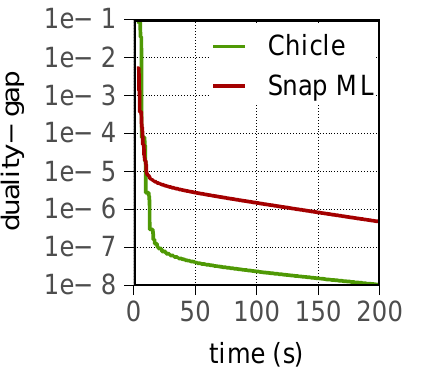}
    \label{fig:evaluation:reference:convergence-per-time:criteo}
  }
  \caption{\ok{Comparison with Snap~ML w.r.t.\ convergence over epochs (top) and time (bottom).}}
  \label{fig:evaluation:reference:convergence-per-epoch}
  \label{fig:evaluation:reference:convergence-per-time}
\end{figure}

\newpage
\subsection{Elastic scaling}
\label{sec:appendix:evaluation:scaling}
\begin{figure*}[t]
  \centering
  \subfloat[CIFAR-10] {%
    \parbox{0.245\linewidth}{
      \includegraphics[width=\linewidth]{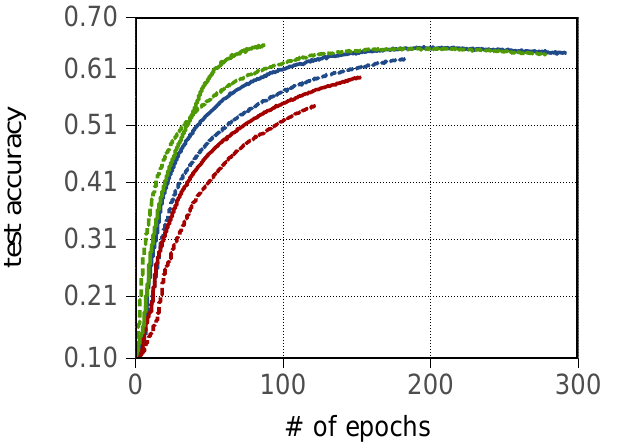}
      \label{fig:evaluation:scaling:convergence-per-epoch:cifar-10}
    }
  }
  \subfloat[Fashion-MNIST] {%
    \parbox{0.245\linewidth}{
      \includegraphics[width=\linewidth]{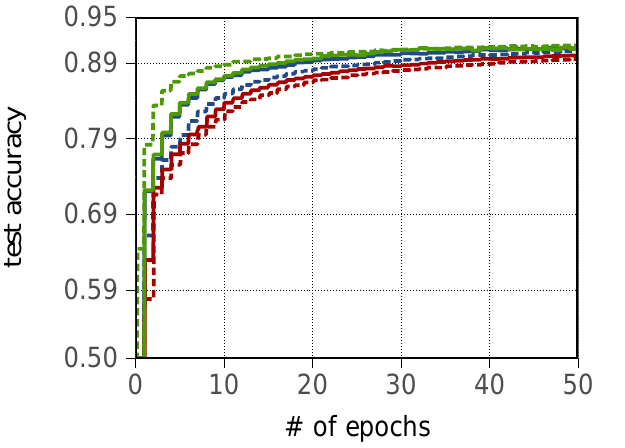}
      \label{fig:evaluation:scaling:convergence-per-epoch:fashion-mnist}
    }
  }
  \subfloat[Higgs] {%
    \parbox{0.245\linewidth}{
      \includegraphics[width=\linewidth]{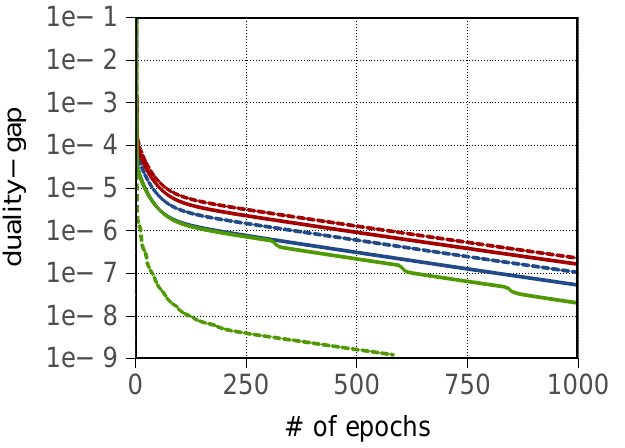}
      \label{fig:evaluation:scaling:convergence-per-epoch:higgs}
    }
  }
  \subfloat[Criteo] {%
    \parbox{0.245\linewidth}{
      \includegraphics[width=\linewidth]{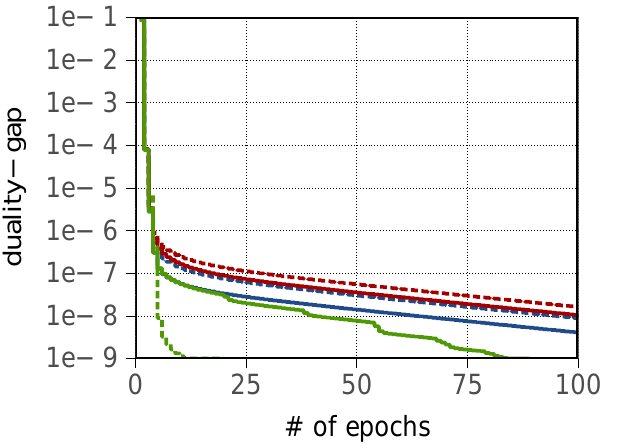}
      \label{fig:evaluation:scaling:convergence-per-epoch:criteo}
    }
  }
  \vspace{0.2cm}
  \includegraphics[scale=1.0]{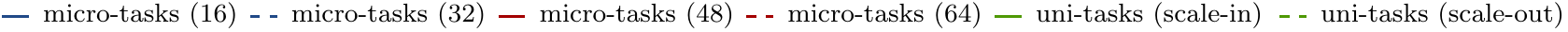}
  \caption{Convergence per epoch vs. data parallelism for elastic scale-in/out
    experiments.  The number of micro-tasks is given in parentheses.}
  \label{fig:evaluation:scaling:convergence-per-epoch}
\end{figure*}

\Fig{evaluation:scaling:convergence-per-epoch} shows per-epoch convergence
results for the elastic scaling experiments.

\subsection{Load balancing}
\label{sec:appendix:evaluation:balance}
\begin{figure*}[t]
  \centering
  \subfloat[CIFAR-10] {%
    \parbox{0.245\linewidth}{
      \includegraphics[width=\linewidth]{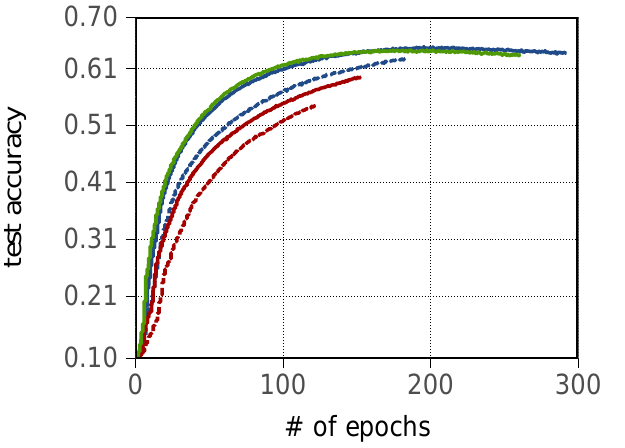}
      \label{fig:evaluation:balance:convergence-per-epoch:cifar-10}
    }
  }
  \subfloat[Fashion-MNIST] {%
    \parbox{0.245\linewidth}{
      \includegraphics[width=\linewidth]{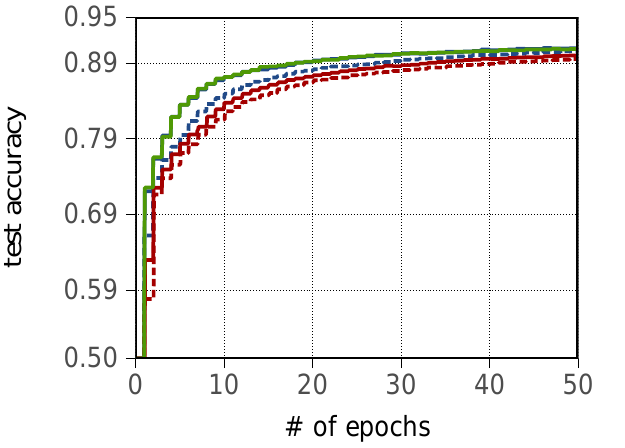}
      \label{fig:evaluation:balance:convergence-per-epoch:fashion-mnist}
    }
  }
  \subfloat[Higgs] {%
    \parbox{0.245\linewidth}{
      \includegraphics[width=\linewidth]{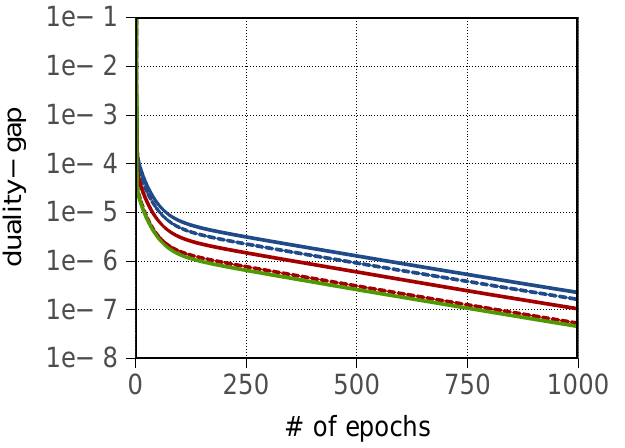}
      \label{fig:evaluation:balance:convergence-per-epoch:higgs}
    }
  }
  \subfloat[Criteo] {%
    \parbox{0.245\linewidth}{
      \includegraphics[width=\linewidth]{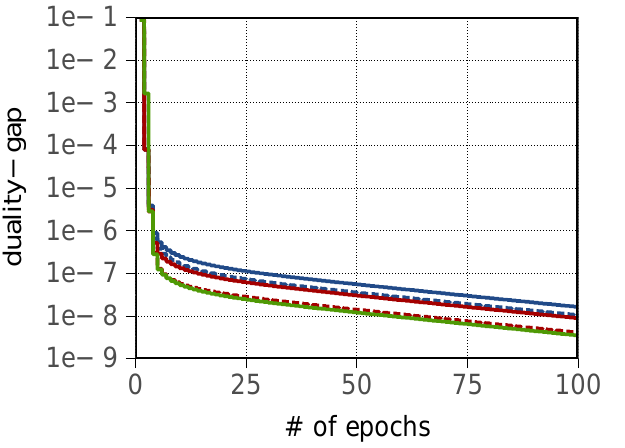}
      \label{fig:evaluation:balance:convergence-per-epoch:criteo}
    }
  }
  \vspace{0.2cm}
  \includegraphics[scale=1.0]{figures/legend}
  \caption{Convergence per epoch when balancing load balancing in a
    heterogeneous cluster.  The number of micro-tasks is given in parentheses.}
  \label{fig:evaluation:balance:convergence-per-epoch}
\end{figure*}

\begin{figure*}[t]
   \centering
   \subfloat[CIFAR-10] {%
     \parbox{0.245\linewidth}{
       \includegraphics[width=\linewidth]{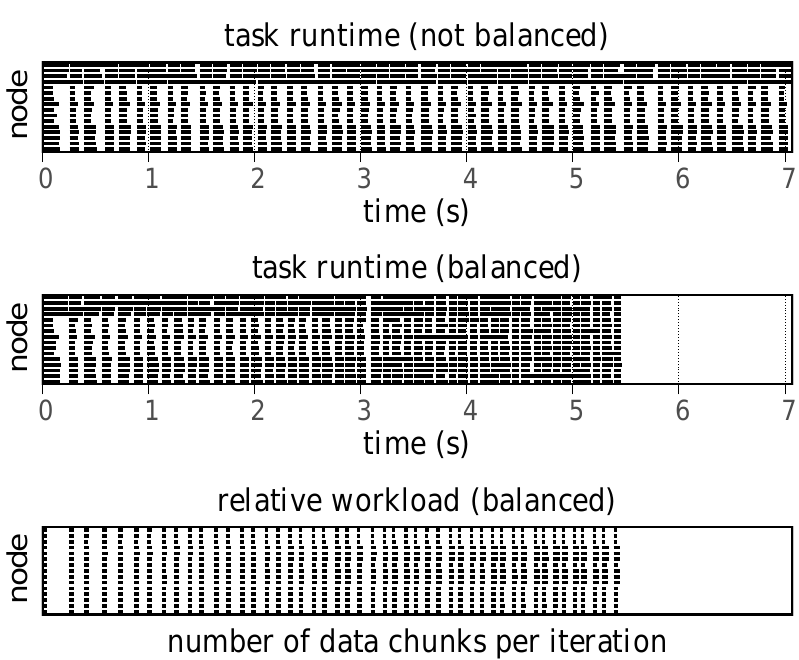}
       \label{fig:evaluation:real:balance:cifar-10}
     }
   }
   \subfloat[Fashion-MNIST] {%
     \parbox{0.245\linewidth}{
       \includegraphics[width=\linewidth]{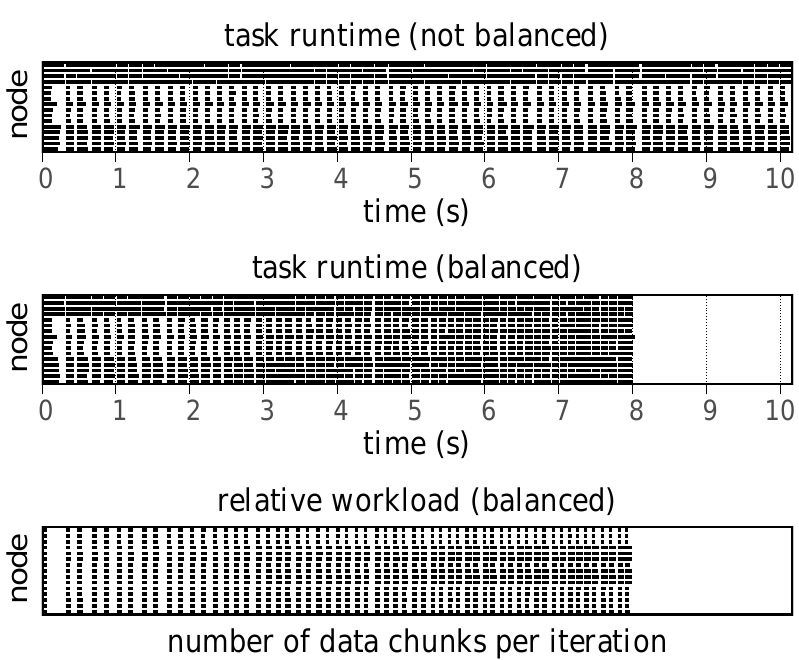}
       \label{fig:evaluation:real:balance:fashion-mnist}
     }
   }
   \subfloat[Higgs] {%
     \parbox{0.245\linewidth}{
       \includegraphics[width=\linewidth]{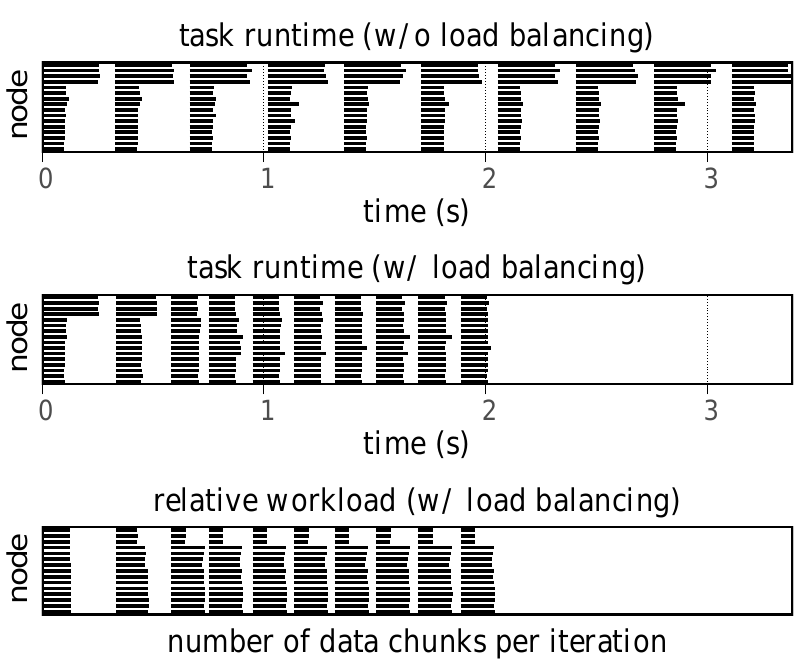}
       \label{fig:evaluation:real:balance:higgs}
     }
   }
   \subfloat[Criteo] {%
     \parbox{0.245\linewidth}{
       \includegraphics[width=\linewidth]{data/cocoa/plot-heterogen-execution-criteo.pdf}
       \label{fig:evaluation:real:balance:criteo}
     }
   }
   \caption{Task execution duration and per worker workload for the load
     balancing in a heterogeneous cluster experiments.}
   \label{fig:evaluation:real:balance}
 \end{figure*}

\Fig{evaluation:balance:convergence-per-epoch} shows per-epoch convergence
results for the load balancing experiments.  \Fig{evaluation:real:balance} shows
the load balancing process during the first 10 (\gls{cocoa}) and 50 (\gls{lsgd})
iterations.

\subsection{Example application}
\label{sec:appendix:example}
\Lst{system:app} shows a simplified trained and solver module for \gls{msgd} on
\sys.

\lstset{language=C++, caption={A minimal \Sys application example},
  label=lst:system:app}
\begin{lstlisting}
  Trainer::run() {
    while (!done) {
      signal(StartIteration);
      wait(IterationFinished);
      model = merge_updates(); // merge and
      broadcast(model);        // broadcast updates
    }
  }

  Solver::run() {
    while (!done) {
      wait(IterationStarted);
      model = get_model() // fetch model
      // perform training
      sample = get_next_sample();
      output = model->forward(sample->data);
      loss = compute_loss(output, sample->label);
      loss->backward();
      sgd_optimizer.step();
      send(model) // post updates
      signal(IterationFinished);
    }
  }
\end{lstlisting}


\end{document}